\documentclass[conference]{IEEEtran}
\IEEEoverridecommandlockouts
\usepackage{cite}
\usepackage{amsmath,amssymb,amsfonts}
\usepackage{algorithmic}
\usepackage{graphicx}
\usepackage{textcomp}
\usepackage{xcolor}
\usepackage{subfigure}
\def\BibTeX{{\rm B\kern-.05em{\sc i\kern-.025em b}\kern-.08em
    T\kern-.1667em\lower.7ex\hbox{E}\kern-.125emX}}

\makeatletter
\newcommand{\linebreakand}{%
\end{@IEEEauthorhalign}
\hfill\mbox{}\par
\mbox{}\hfill\begin{@IEEEauthorhalign}
}
\makeatother

\begin{document}

\title{A Graph Neural Network with Negative Message Passing for Graph Coloring
\thanks{
This work was supported by an Alexander von Humboldt Professorship for Artificial Intelligence endowed to Y. Jin by the German Federal Ministry of Education and Research.
}
}

\author{\IEEEauthorblockN{1\textsuperscript{st} Xiangyu Wang}
\IEEEauthorblockA{\textit{Faculty of Technology} \\
\textit{Bielefeld University}\\
33619 Bielefeld, Germany \\
0000-0002-1832-0905}
\and
\IEEEauthorblockN{2\textsuperscript{nd} Xueming Yan}
\IEEEauthorblockA{\textit{School of Information Science and Technology; Faculty of Technology} \\
\textit{Guangdong University of Foreign Studies; Bielefeld University}\\
Guangzhou 510000, China; 33619 Bielefeld, Germany \\
yanxm@gdufs.edu.cn}
\linebreakand
\IEEEauthorblockN{3\textsuperscript{rd} Yaochu Jin $^{\ast}$ \thanks{*Corresponding author}}
\IEEEauthorblockA{\textit{Faculty of Technology} \\
	\textit{Bielefeld University}\\
	33619 Bielefeld, Germany \\
	yaochu.jin@uni-bielefeld.de}
}


\maketitle

\begin{abstract}
Graph neural networks have received increased attention over the past years due to their promising ability to handle graph-structured data, which can be found in many real-world problems such as recommended systems and drug synthesis. Most existing research focuses on using graph neural networks to solve homophilous problems, but little attention has been paid to heterophily-type problems. In this paper, we propose a graph network model for graph coloring, which is a class of representative heterophilous problems. Different from the conventional graph networks, we introduce \textit{negative message passing} into the proposed graph neural network for more effective information exchange in handling graph coloring problems. Moreover, a new loss function taking into account the self-information of the nodes is suggested to accelerate the learning process. Experimental studies are carried out to compare the proposed graph model with five state-of-the-art algorithms on ten publicly available graph coloring problems and one real-world application. Numerical results demonstrate the effectiveness of the proposed graph neural network.
\end{abstract}


\section{Introduction}

Different types of data, such as traditional unordered data, time series structured data, and graph-structured data may be encountered in solving real-world optimization problems. Graph-structured data contains rich relationship information between the attributes, making it more challenging to effectively learn the knowledge in the data using traditional machine learning models. Recently, graph neural networks (GNNs) have become extremely popular in the machine learning community, thanks to their strong ability to capture relational information in the data. Many different types of GNNs have been proposed, and they can be usually categorized according to different aggregation methods, such as graph convolutional network (GCN) \cite{GCN}, GraphSAGE \cite{GraphSAGE}, graph attention network (GAT) \cite{GAT}, among others. Since GNN enables the nodes to aggregate their neighbors' information, it becomes a powerful tool for solving problems in social networking \cite{social}, bioinformatics \cite{bio}, and community detection \cite{detection}, to name a few.  Node classification is often considered as bi- or multi-class classification problems by inputting the features of nodes and edges and outputting each node's probability belonging to different classes. During the optimization process, each node combines information of its own and its neighbors to generate embedding vectors. By representing node and edge features in a higher dimensional space, embedding vectors are used to solve problems by downstream machine learning tasks in a non-autoregression or autoregression way \cite{survey}.

However, most of the optimization problems mentioned above are continuous and homophilous. That is, two nodes between one edge tend to be very similar, i.e., have similar embedding vectors. These homophilous problems described as graphs have many real-world applications, and have been studied extensively \cite{rec1,rec2,prot}. On the contrary, some problems contain heterophily property \cite{hetro} where the connected nodes have as different embeddings as possible. For example, graph coloring problems (GCPs) are a type of classical heterophilous problems because they are defined to have different colors between the connected nodes. GCPs have attracted increased research interest since many real-world applications can be formulated as graph coloring problems, such as arranging timetables and schedules, managing air traffic flow, and many others \cite{app-s}. In fact, GCPs, like most combinatorial optimization problems (COPs), are NP-hard problems, which require highly intensive computational costs to obtain the exact optimal solutions. Luckily, some approaches have been proposed to find approximate optimal solutions of COPs with the help of GNNs within an acceptable period of computation time \cite{TSP, CO-GNN, GNN-CO}.  For example, Liu \textit{et al.} \cite{Liu2023EMO} proposed two supervised residual gated GCNs to directly predict the entire Pareto set for multi-objective facility location problems.

\begin{figure*}[ht]
	\centering
	\subfigure[]{
		\centering
		\includegraphics[width=0.25\linewidth]{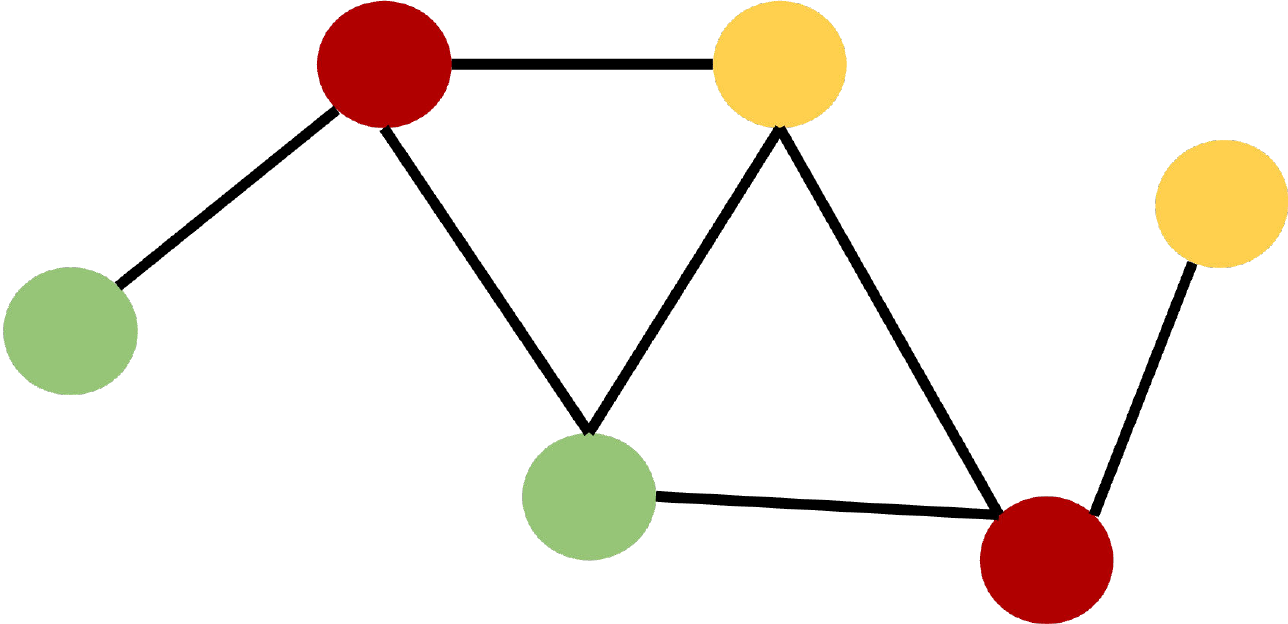}
		\hspace{8mm}
		
	}%
	\subfigure[]{
		\centering
		\includegraphics[width=0.25\linewidth]{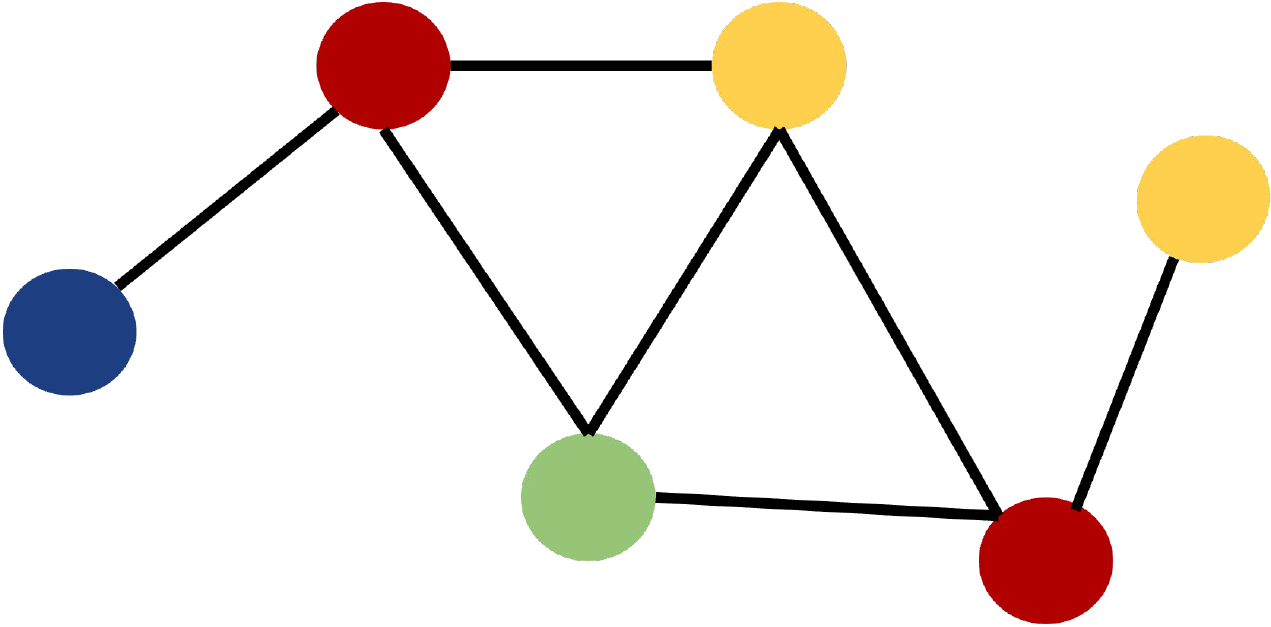}
		\hspace{8mm}
		
	}%
	\subfigure[]{
		\centering
		\includegraphics[width=0.25\linewidth]{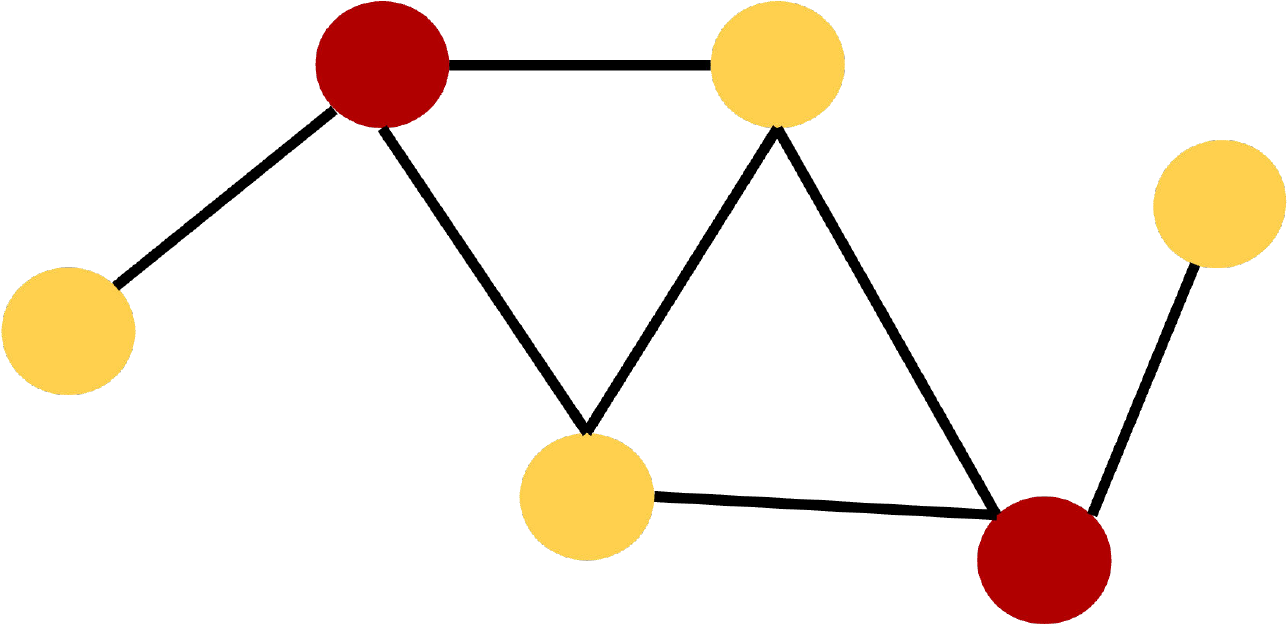}
		\hspace{8mm}
	}%
	\centering
	\caption{\label{fig1}Example solutions to graph coloring with six nodes and seven edges. (a) An optimal solution for the given problem, where $\mathcal{X}(\mathcal{G})=3$. (b) The number of total conflicts is zero, but the number of used colors can be reduced. (c) A solution using only two colors, while the $\mathcal{X}(\mathcal{G})$ being $3$, and the number of total conflicts should be minimized.}
\end{figure*}

GNNs have inherent advantages to represent GCPs and solve such graph-structured problems, as samples (nodes) can aggregate information from their neighbors. Recently, some efforts have been made to apply GNNs to solve GCPs, and different structured graph neural networks have been proposed. GNN-GCP \cite{GNN-GCP} uses a network containing a GNN to predict the chromatic number of a given graph and uses the generated embedding vectors for clustering to obtain a color assignment. Li \textit{et al.} \cite{GDN} explored several rules for using a GNN to solve GCPs, and proposed a graph neural network called GDN. PI-GNN \cite{PI-GNN} is inspired by a Potts model, and it is shown to perform well by combining with a novel loss function.

The above algorithms basically follow the classical network structures, which are proposed to solve homophilous problems, such as node classification and link prediction \cite{lp}. However, these structures may not necessarily be suited for solving GCPs, and improvements should be made in the structural design to better adapt to the heterogeneous characteristics of GCPs. For this reason, a negative message passing strategy is proposed considering the requirements for GCPs, i.e., two connected nodes should have as different embedding vectors as possible. Meanwhile, color assignment with no conflict and stable convergence during the training process are both desired when solving GCPs. Therefore, a loss function consisting of a utility objective and a convergence objective is proposed to guarantee the good performance of the proposed neural network and a stable training process.

Section II presents the preliminaries of this work, including the definition of graph coloring problems, the aggregation methods of classical GNNs, and the related work that use GNNs to solve graph coloring problems. In Section III, the proposed graph network model with negative message passing and a new loss function is detailed. Section IV describes the experimental results on the benchmark graph coloring problems and a real-world problem. Ablation studies and analysis of the computational complexity are also presented. Finally, conclusions and future work are given in Section V.

\section{Preliminaries}
\subsection{Definition of Graph Coloring Problems}

We consider an undirected graph $\mathcal{G}=(\mathcal{V}, \mathcal{E})$, where $\mathcal{V}=\{1,2,\dots, n\}$ is the node set, and $\mathcal{E}$ is the edge set connecting two nodes, represented by $(u,v)\in \mathcal{E}$. The adjacency matrix $A$ of a graph $\mathcal{G}$ also represents the connection between the nodes. When nodes $i$ and $j$ are connected, $A(i,j)$ and $A(j, i)$ are equal to 1; otherwise, elements in these two positions of $A$ are equal to 0. 

Graph coloring problems \cite{GCP1, GCP2, GCP3} aim to minimize the number of conflicts while using the minimum number of colors. Mathematically, GCPs can be formulated in two different forms. In the first formulation, GCPs are defined as a constrained minimization problem as follows:
\begin{equation}
	\label{M1}
	\begin{split}
		&\min \quad k,\\
		&\mathrm{s.t. } \sum_{(u,v)\in \mathcal{E}}f(u,v)=0,
	\end{split}
\end{equation}  
where $f(u,v)$ represents the clash between neighboring nodes $u$ and $v$, and $(u,v)\in \mathcal{E}$. If $u$ and $v$ share the same color, $f(u,v)=1$; otherwise, $f(u,v)=0$. $k$ is the number of colors used in graphs. If $k$ colors can fill a graph with no clash, this graph is called $k$-colorable. The chromatic number $\mathcal{X}(\mathcal{G})$ is the minimum of $k$, denoting the optimum number of colors without resulting in any conflicts in $\mathcal{G}$.

Alternatively, GCPs can also be expressed as a constraint satisfaction problem with a given number of colors $k$:
\begin{equation}
	\label{M2}
	\begin{split}
		&\min \sum_{(u,v)\in \mathcal{E}}f(u,v),\\
		&\mathrm{s.t.} \quad   k .
	\end{split}
\end{equation}  
This formulation aims to minimize the clashes in a graph, given that the number of colors that can be used is $k$.

The optimization problems defined in Eq. (\ref{M1}) and Eq. (\ref{M2}) are slightly different, which may be suited for different requirements or priorities in solving the same problem. Take the assignment of taxis to customer requests as an example. In Eq. (\ref{M1}), the customer requests are the top priority, and a minimal number of taxis should be found. In this case, the assumption is that there is a sufficient number of taxis. However, if the number of taxis is limited during the peak period, the assignment that can satisfy the majority of customer requests would be preferred. This work will focus on minimizing the number of total conflicts under a certain given number of colors, i.e., the GC problem will be solved according to the formulation in Eq. (\ref{M2}).

An illustrative example consisting of six nodes and seven edges is given in Fig. \ref{fig1}. In Fig. \ref{fig1} (a), three colors are used, and no connected neighboring nodes have the same color, meaning that this is an optimal solution to the given example. By contrast, as shown in Fig. \ref{fig1}(b), there is no conflict between the neighboring nodes, but the number of colors used can still be further minimized. In Fig. \ref{fig1} (c), on the other hand, only two colors have been used, and there is one conflict in color. Overall, Fig. \ref{fig1} (a) provides an ideal solution, while solutions in Fig. \ref{fig1} (b) and (c) may also be applicable in different scenarios. 

\subsection{Classical GNNs}
The main difference between graph neural networks and other neural networks is that the nodes (samples) in GNNs can gather information from their neighbors before being projected to the next layer, due to the existence of edges \cite{AG1, AG2}. Aggregating embeddings from neighbors and combining them with the node's embedding are two critical operators in GNNs called aggregation and combination. Authors in \cite{GDN} consider GNNs with aggregation and combination operators as AC-GNNs. They make GNNs powerful in exploiting and revealing relationships between nodes.

The input to GNNs is usually the feature vector or randomly-generated vector of each node. With an adjacency matrix, one node generates a new embedding in the next hidden layer by combining its own and neighbors' aggregated embedding. Therefore, embeddings in the first layer contain first-order neighborhood information and the $k$-th hidden layer captures the $k$-th neighborhood information. The aggregation and combination operators may differ in various GNNs based on different purposes and optimization tasks. In the following, we briefly review some classical and popular aggregation and combination methods.

GCN \cite{GCN} considers the equal importance of a node and its neighbor information, which integrates the aggregation and combination methods. The embedding of the $k$-th layer is calculated by $H^k=\sigma (\hat{A}H^{k-1}W^{k-1})$, where $\hat{A}=\tilde{D}^{-\frac{1}{2}}\tilde{A}\tilde{D}^{-\frac{1}{2}}$ is the normalized adjacency matrix with self connections, $D$ is the degree matrix, and $W^{k-1}$ is a trainable weight matrix of the $k-1$-th layers. On the other hand, some GNNs separate the aggregation and combination methods, assigning different importance to the node embedding and the aggregated neighborhood embedding. GraphSAGE \cite{GraphSAGE} is one example of this type; the embedding of node $v$ in the $k$-th hidden layer is obtained by $h_v^k\gets \sigma (W^{k-1}\cdot CONCAT(h_v^{k-1},h_{\mathcal{N}(v)}^k))$, where $h_{\mathcal{N}(v)}^k\gets AGGREGATE(h^{k-1}_{u\in \mathcal{N}(v)})$. There are also many other types of aggregation and combination methods proposed very recently, such as HetGNN \cite{HetG}, DNA \cite{DNA}, non-local graph neural networks \cite{NL}, and SAR \cite{SAR}. 


\subsection{Related Work}
Only sporadic work on solving graph coloring problems with the help of graph neural networks have been reported, which are based on supervised or unsupervised learning.

GNN-GCP \cite{GNN-GCP} generates color embeddings and node embeddings with a process of learning whether the graph is $k$-colorable or not. At first, $2^{15}$ positive and $2^{15}$ negative GCP instances are generated with ground truth given by a GCP solver. The proposed neural network learns the difference between the ground truth and the prediction by using the binary cross entropy as a loss function. The prediction is gotten through a GNN to aggregate neighbor information, an RNN to update the embeddings, and an MLP to get the final logit probability. If the network predicts a graph is k-colorable, the second stage is applied by clustering vertex embeddings using k-means, and nodes in the same cluster share the same color. 

Different from the above work that relies on supervised learning and a set of pre-solved GCP instances, unsupervised learning has also been adopted to tackle GCPs by constructing a loss function without requiring the ground truth. In general, the output of unsupervised learning for GCPs are probability vectors of the nodes, indicating which color should be assigned to each node. GDN \cite{GDN} uses a margin loss function, which minimizes the distance between a pre-defined margin and a Euclidean distance between node pairs. Schuetz \textit{et al.} \cite{PI-GNN} take advantage of the close relationship between GCPs and the Potts model so that the partition function of the Potts model can be converted to the chromatic function of $\mathcal{G}$ mathematically. Besides, the Potts model only distinguishes whether neighboring spins are in the same state or not, which is very similar to the definition of GCPs. Consequently, a loss function is proposed in PI-GNN that minimizes the inner product of node pairs. Generally, loss functions in unsupervised learning usually aim to minimize the similarity between the connected nodes in solving heterophilous problems such as graph coloring problems. 

Besides, solving GCPs with heuristic algorithms has also been studied. Tabucol \cite{Tabucol} is a Tabu-based algorithm that defines Tabu moves and a Tabu list tailored for GCPs. It changes the color (for example, red) of one randomly selected node into another color (green) to reduce the number of conflicts. The color set (green, red) will be put into the Tabu list for certain iterations to restrict this node from changing to red. An evolutionary algorithm, HybridEA \cite{HyridEA} is proposed to modify traditional ways of generating offspring and use a Tabu-search method as the mutation operator. 

\section{The Proposed Graph Neural Network Model}

The main difference between heterophilous and homophilous optimization problems is that connected nodes in shared edges should express differently rather than similarly. For example, in node classification tasks, neighboring nodes should be classified into different classes instead of the same class. Therefore, the embeddings of connected nodes should be as different as possible, and the first-order neighbors need to pass negative messages to the node. On the other hand, the second-order neighborhood may contain information that positively impacts the node. Therefore, inspired by the heterophilous property of graph coloring problems, we proposed a new AC-GNN, which is called GNN-1N by mainly focusing on the first-order (1st) negative message passing strategy.


To better solve GCPs, the proposed algorithm should take into account two objectives. One is to find a solution without conflict and the other is to achieve fast and stable convergence. Based on the above two objectives, a loss function for solving graph coloring problems in an unsupervised way is proposed.

\subsection{Forward Propagation}
The forward propagation of the proposed framework is described in this subsection. We apply GraphSAGE \cite{GraphSAGE} as a baseline GNN model, and some modifications specialized for solving GCPs are made. In the original paper of GraphSAGE, the authors give three aggregation methods: mean aggregator, LSTM aggregator, and pool aggregator. The embedding $h_v^k$ of node $v$ in the $k$-th layer using the mean aggregator is calculated as follows:
\begin{equation}
	\label{G-mean}
	\begin{split}
		&h_{\mathcal{N}(v)}^k=mean\{h_u^{k-1}\}, \forall u\in \mathcal{N}(v),\\
		&h_v^k = \sigma\left(W_{self}^k\cdot h_v^{k-1} + W_{neigh}^k \cdot h^k_{\mathcal{N}(v)} \right),
	\end{split}
\end{equation} 
where $\mathcal{N}(v)$ is the connected node set of node $v$, $h_{\mathcal{N}(v)}^k$ is the aggregated embedding of neighborhood of $v$, $\sigma$ is the activation operator, $W_{self}^k$ and $W_{neigh}^k$ is the learnable weight matrix for the $k$-th layer. In this work, the mean aggregator is considered for improvement.

At the beginning of forward propagation, the embedding of each node is usually the feature vector. However, some graphs in GCPs may not have features, so embeddings are generated randomly. In the first hidden layer, nodes gather the first-order neighborhood information, which should make negative contributions. Thus, the embedding of the first layer is obtained by 
\begin{equation}
	\label{MyG}
	\begin{split}
		&h_{\mathcal{N}(v)}^1=mean\{h_u^{0}\}, \forall u\in \mathcal{N}(v),\\
		&h_v^1 = \sigma\left(W_{self}^1\cdot h_v^{0} - \alpha \cdot W_{neigh}^1 \cdot h^1_{\mathcal{N}(v)} \right),
	\end{split}
\end{equation} 
where $h_v^0$ is the randomly generated feature of the input layer of node $v$, $\alpha$ is the trainable parameter controlling the negative influence of the neighborhood. Elements in  $W_{self}^1$ and $W_{neigh}^1$ are all positive values distributed uniformly from 0 to 1, and $\alpha$ is initialized to be $0.5$.

This strategy is reasonable and natural. For example, we assume that there is node $v$ and its two neighbors $u_1$ and $u_2$, whose embeddings are $[0.8,0.6,0.1]$, $[0.7,0.1,0.1]$, and $[0.5,0.1,0.7]$, respectively. We let $W_{self}^1$ and $W_{neigh}^1$ be an identity matrix, and $\alpha=0.5$, the embedding of node $v$ is $[0.5, 0.55, -0.1]$ calculated by Eq. (\ref{MyG}). If embeddings also represent the probability of the assigned color, $v$ and $u_1$ conflict with each other before applying Eq. (\ref{MyG}), as these two nodes both prefer the first color among the three given colors. The assigned color of node $v$ changes into the second one after applying Eq. (\ref{MyG}). On the other hand, if Eq. (\ref{G-mean}) is used as an updating strategy in the first hidden layer, $v$ and $u_1$ will remain to be conflicting with each other. 

According to the property of GNNs, the second hidden layer can aggregate the second-order neighborhood information, which may be able to contribute some helpful positive influence. Therefore, in the second-order layer, the original mean aggregator (Eq. (\ref{G-mean})) of GraphSAGE is used to generate embeddings $h_v^2$.

For graph coloring problems, there are two ways to assign colors to nodes. Firstly, there are $k$ sets containing nodes, and all nodes in one set have the same color. The second way uses the $k$-length probability vector of each node, and the node is assigned to the $i$-th color if the probability in the $i$-th position of the vector is the largest. This work considers GCPs as an unsupervised classification task, and the second way is used to assign $k$ colors to different nodes. Therefore, the dimension of $h_v^2$ equals the color number $k$. The probability of node $v$ can be obtained as follows: 
\begin{equation}
	\label{p}
	p_v=softmax\{h_v^2\},
\end{equation} 
where $softmax$ is the softmax operator, i.e., $p_v(j)=\frac{h_v^2(j)}{\sum_{i=1}^{k}h_v^2(i)}$. The final color assigned to $v$ is the color with the highest probability. 

\subsection{Loss Function}
A loss function is proposed to achieve the utility-based and the convergence-based goal in an unsupervised way. The former is to minimize the conflicts between the connected nodes with a given number of colors, while the latter is to minimize the uncertainty of nodes, which can stabilize convergence. 

As no ground truth is available, the utility-based objective function uses the probability of nodes to reflect the relationship of connected nodes. The main idea is to maximize the difference or minimize the similarity between node pairs. Some loss functions have been proposed based on the above idea. Among them, the loss function inspired by the Potts model performs very well, which is therefore adopted as the utility-based objective function in this work:
\begin{equation}
\label{f_u}
	f_{utility}=\sum_{(u,v)\in \mathcal{E}}p_v^T\cdot p_u.
\end{equation} 
Eq. (\ref{f_u}) only aims to minimize the inner product between two probabilities, which can be considered as minimizing the similarity between two nodes. This is reasonable for solving heterophilous problems such as GCPs. Note that no other \textit{a priori} knowledge is required, making it suitable for solving other unsupervised learning problems.
\begin{figure}[ht]
	
	\centering
	\includegraphics[width=1\linewidth]{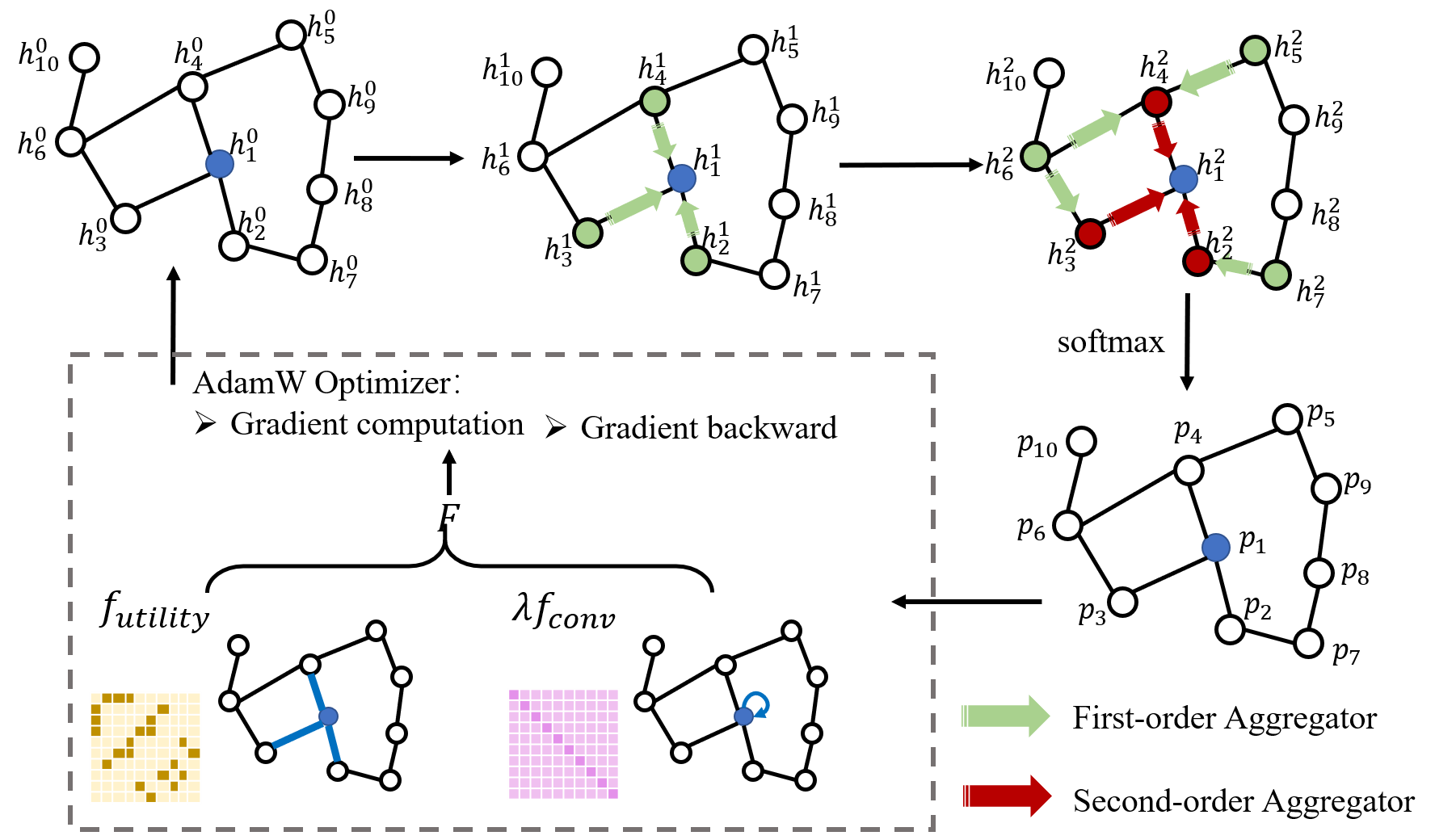}
	\caption{\label{fig2}The framework of the proposed method, where the first-order aggregator applies Eq. (\ref{MyG}) and the second-order aggregator uses the original aggregator Eq. (\ref{G-mean}). Between the first and the second hidden layers, dropout is employed to avoid getting stuck in local optimums.}
\end{figure}

In addition to maximizing the difference between the connected nodes, we also intend to increase the probability of one node being assigned a certain color to make the learning process more stable. Therefore, we introduce self-information to construct $f_{conv}$, which is formulated as follows:
\begin{equation}
	\label{f_c}
	f_{conv}=\sum_{i=1}^n p_i^T\cdot \log p_i.
\end{equation} 
Self-information represents the amount of information in an event. If the value of self-information is large, it means this event contains more information, indicating a high uncertainty. Otherwise, the uncertainty of this event is low. In terms of GCPs, high self-information of one node means the probabilities of being assigned to different colors are approximately equal. It results in unstable convergence because the color assignment varies greatly under small weight changes. Therefore, nodes with small self-information hold more confidence in the current color assignments, which helps stabilize the convergence.

\begin{table*}
	\caption{Numerical results and data information for the COLOR dataset.}
	\label{table1}
	\begin{tabular}{c|cc|c|cccccc}
		\hline
		\hline
		Graph Name &\# Nodes&\# Edges&Color Number & Tabucol \cite{Tabucol}& HybridEA \cite{HyridEA} & GDN \cite{GDN} & PI-GCN \cite{PI-GNN}& PI-SAGE \cite{PI-GNN}&GNN-1N\\
		\hline
		myciel5 & 47& 236&6&\pmb{0}&\pmb{0}&\pmb{0}&\pmb{0}&\pmb{0}&\pmb{0} \\
		myciel6& 95& 755&7 &\pmb{0}&\pmb{0}&\pmb{0}&\pmb{0}&\pmb{0}&\pmb{0}\\
		queen5-5& 25& 160&7 &\pmb{0}&\pmb{0}&\pmb{0}&\pmb{0}&\pmb{0}&\pmb{0}\\
		queen6-6& 36& 290&7 &\pmb{0}&\pmb{0}&4&1&\pmb{0}&\pmb{0}\\
		queen7-7& 49& 476&7 &10&9&15&8&\pmb{0}&\pmb{0}\\
		queen8-8& 64& 728&9 &8&5&7&6&\pmb{1}&\pmb{1}\\
		queen9-9& 81& 1056&10&5 &6&13&13&\pmb{1}&\pmb{1}\\
		queen8-12& 96& 1368&12 &10&3&7&10&\pmb{0}&\pmb{0}\\
		queen11-11& 121& 1980&11 &33&22&33&37&17&\pmb{13}\\
		queen13-13& 169& 3328&13 &42&37&40&61&26&\pmb{15}\\
		
		\hline
	\end{tabular}
\end{table*}
By combining the terms in Eq. (\ref{f_u}) and Eq. (\ref{f_c}), we get the following loss function:
\begin{equation}
	\label{F}
	\mathop{\min} F=f_{utility}+\lambda f_{conv}.
\end{equation} 
where $\lambda>0$ is a hyperparameter. 

\subsection{The Overall Framework}

As shown in Fig. \ref{fig2}, the framework mainly consists of the forward propagation and the optimization process. In the forward propagation, the randomly generated embedding $h_i^0$ is assigned to the $i$-th node. If we take $n_1$ as an example, it aggregates its first-order negative neighborhood embedding in the first hidden layer, and its second-order neighborhood embedding in the second layer. Between these two hidden layers, the dropout \cite{dropout} is applied to prevent the solver from getting stuck in a local optimum. After $n_1$ aggregates the two-hop neighborhood information, $h_1^2$ is obtained, followed by the softmax function to get a probability vector $p_1$. $p_1$ contains $k$ elements representing the probabilities of choosing $k$ colors. After getting the probability vectors of nodes, the loss function $F$ is calculated with two terms, namely $f_{utility}$ and $f_{conv}$. 
The AdamW optimizer is used to compute the gradient and update weights in the graph neural network.

\section{Numerical Experiments}
In this section, experiments on the COLOR dataset \cite{COLOR} are presented at first, and then an application to taxi scheduling is applied. An ablation study is given to demonstrate the stabilization ability of the proposed loss function, and finally, the computational complexity is analyzed.

\subsection{Experiments on COLOR Dataset}

In this experiment, we use the publicly available COLOR dataset to evaluate the performance of the proposed algorithm and its peer methods. The COLOR dataset is a classical, widely used graph dataset in the field of graph coloring problems, where Myciel graphs are based on the Mycielski transformation and Queens graphs are constructed on $n$ by $n$ chessboard with $n^2$ nodes. More detailed information on graphs can be found in Table 1, with the number of nodes and edges in each graph. Eq. (\ref{M2}) is optimized in the following experiments given the color number $k$, which is shown in Table 1.

Five algorithms are chosen as peer algorithms: Tabucol \cite{Tabucol}, HybridEA \cite{HyridEA}, GDN \cite{GDN}, PI-GCN \cite{PI-GNN}, and PI-SAGE \cite{PI-GNN}. Tabucol and HybridEA are tabu-based heuristics algorithms. The rest three algorithms, GDN, PI-GCN, and PI-SAGE, are GNN-based unsupervised algorithms. The above five methods focus on minimizing conflicts with given numbers of colors. GNN-GCP mainly predicts the color number of a given graph. Therefore, it is not included in this comparison. The results in Table \ref{table1} show the conflicts of each graph ($f(u,v)$ in Eq. (\ref{f_u})) found by the proposed GNN-1N and the algorithms under comparison. 

The results of Tabucol and HybridEA are taken from \cite{GDN}, with a maximum run time, i.e., 24 hours per graph, and the results of GDN, PI-GCN, and PI-SAGE are taken from \cite{PI-GNN}. For a fair comparison, we use the same maximum number of iterations ($10^5$) to run our methods (GNN-1N) on GPU. Besides, the early stopping mechanism is applied within $10^3$ iterations if the value of the loss function changes less than $0.001$. The hyperparameters, including the dimensions of $h_i^k$, the learning rate $\eta$ in the AdamW optimizer, the probability of dropout, and $\lambda$ in the loss function $F$ are optimized in a similar way to that in \cite{PI-GNN}. All results listed in Table \ref{table1} are the best coloring results of all methods, from which we can see that the proposed method performs the best on all graphs, especially on large and dense graphs, such as queen11-11 and queen13-13. On the contrary, traditional tabu-based methods and other machine-learning methods cannot find as few conflicts as GNN-1N does.

\begin{figure}[ht]
	\centering
	\includegraphics[width=0.95\linewidth]{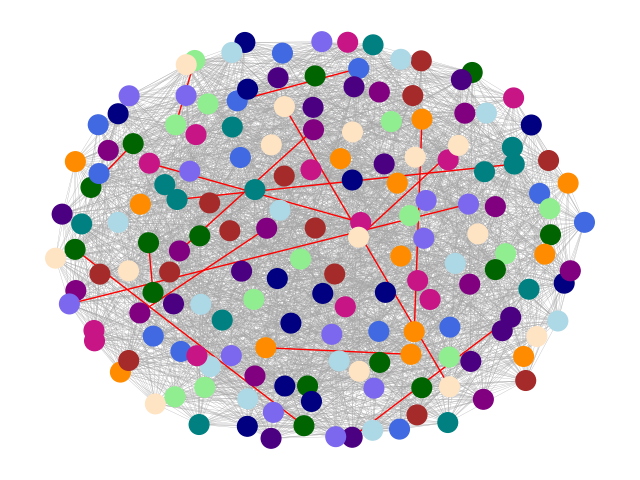}
	\caption{\label{fig3}The final color assignment of queen13-13 with 169 nodes given by our method. Only 15 conflicts highlighted in red lines exist out of 3328 edges (in grey lines) in this figure.}
\end{figure}

\begin{figure*}[ht]
	\centering
	\subfigure[]{
		\centering
		\includegraphics[width=0.27\linewidth]{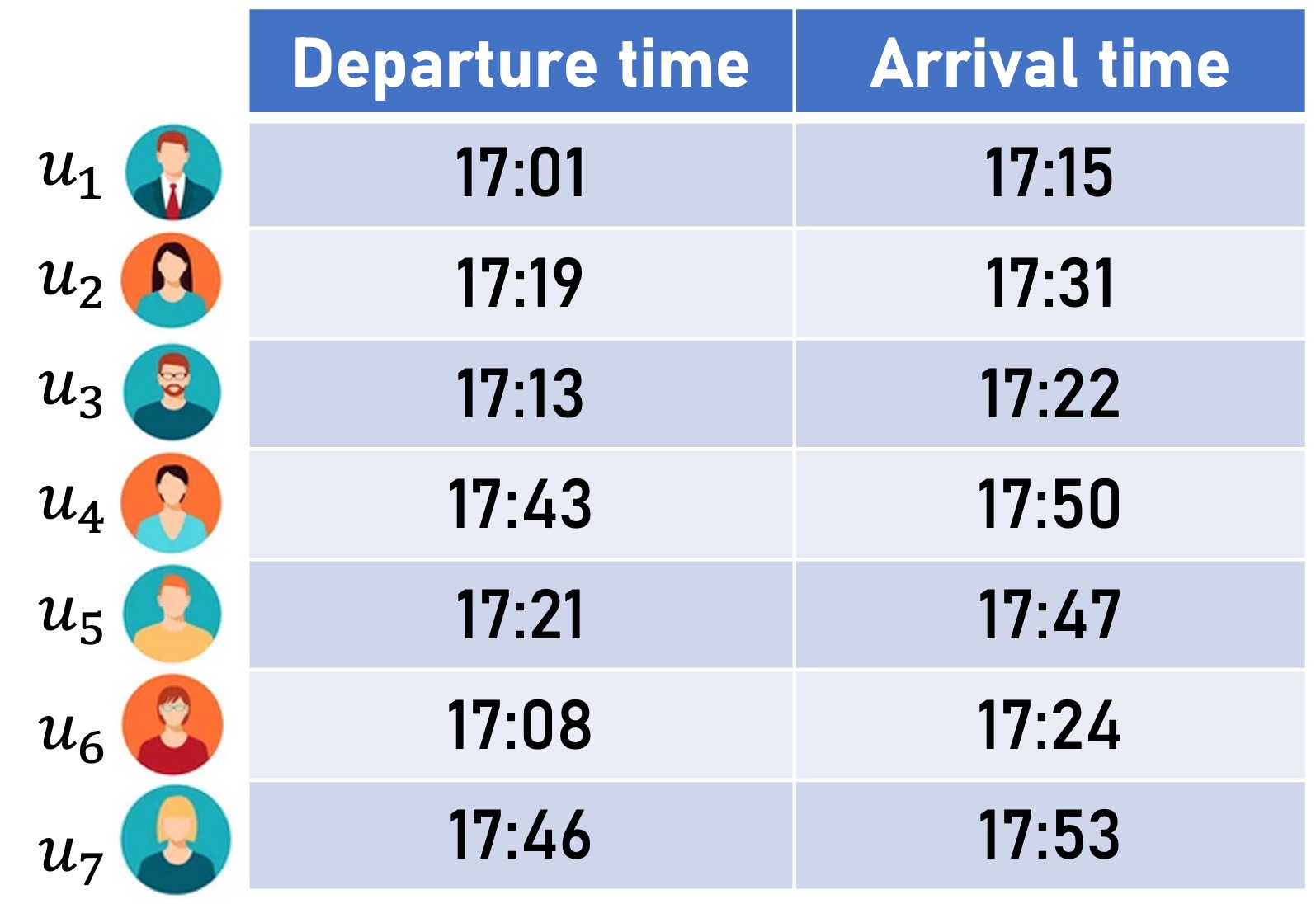}
		\hspace{10mm}
		\label{app1}
	}%
	\subfigure[]{
		\centering
		\includegraphics[width=0.25\linewidth]{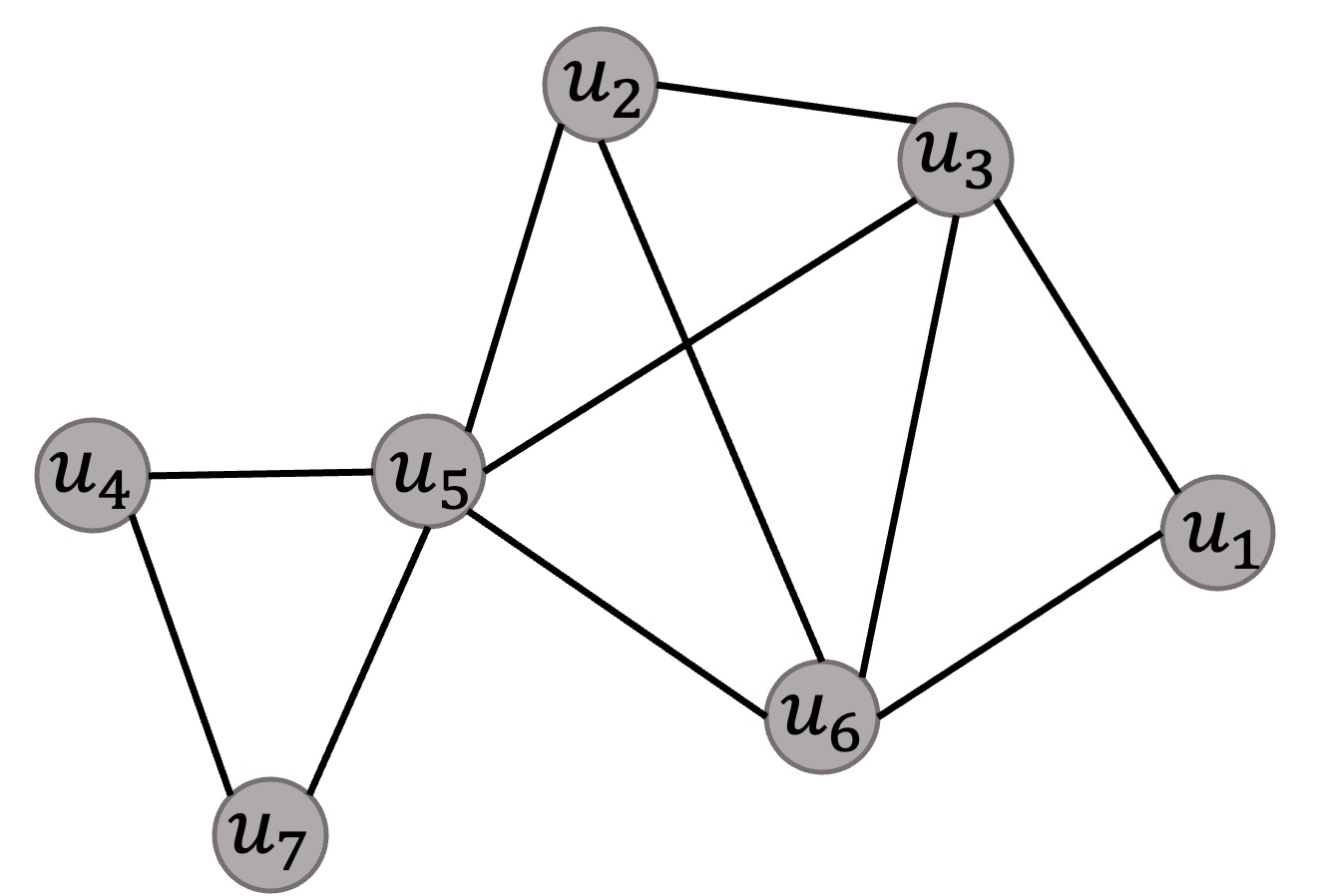}
		\hspace{10mm}
		\label{app2}
	}%
	\subfigure[]{
		\centering
		\includegraphics[width=0.3\linewidth]{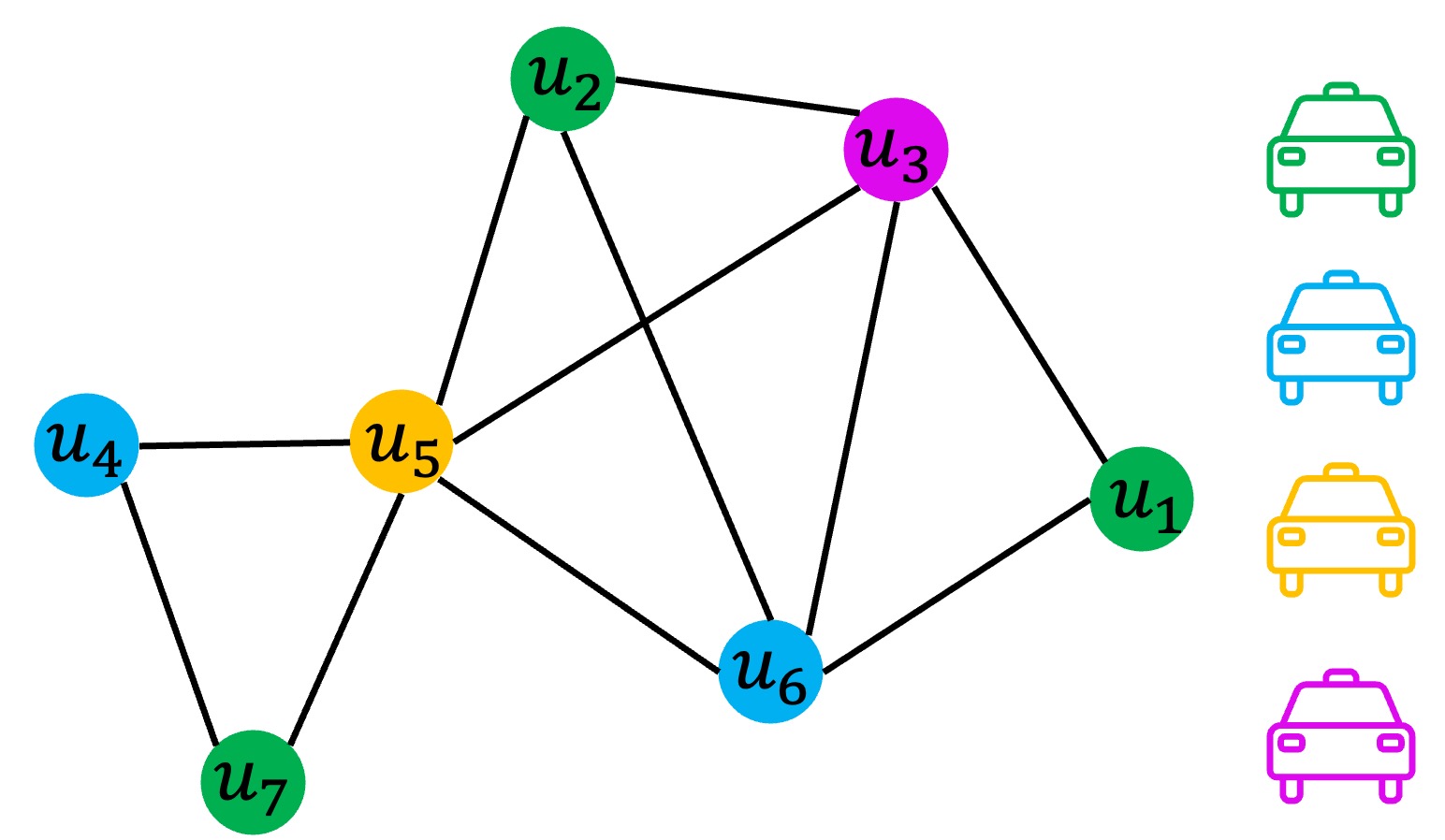}
		\hspace{10mm}
		\label{app3}
	}%
	\centering
	\caption{\label{app}Taxi scheduling problem. (a) The timetable containing departure time and arrival time of seven customers. (b) Encoding the timetable into a graph coloring problem. (c) Optimizing the graph coloring problem with our unsupervised neural network method and decoding it. Under this scenario, customs and taxis are nodes and colors in graph coloring problems, respectively.}
\end{figure*}

To gain more insights into the solutions found by our methods, we plot the color assignment of queen13-13 in Fig. \ref{fig3}. There are 169 nodes in this graph, and 13 colors should be used to color nodes. There are 3328 edges plotted with grey lines and 15 conflicted edges highlighted with red lines. The normalized error rate is $0.45\%=\frac{15}{3328}\times 100\%$, which is relatively small, indicating that our method has a good ability to find less-conflict color assignments.

\subsection{Application}
In this section, we take the taxi scheduling problem as an example to show the ability of GNN-1N to solve graph-structured problems in real life. We give a simple scenario for taxis to customer requests. Seven customs call a taxi company to book taxis one day. They all plan to book a taxi during the evening rush hour from 17:00 to 18:00 and each confirms a time period. The task is to  satisfy all customers' requests with an available number of taxis, assuming that only four taxis are available during this peak hour.

To solve this problem, three steps are taken, which are 1) encoding, 2) optimization, and 3) decoding. The encoding step transfers the given timetable into a graph. As shown in Fig. \ref{app} (a), the timetable with departure time and arrival time of seven customers is shown, and each customer is represented by a node in the graph. As customers with overlapping schedules cannot use the same taxi, two nodes should share one edge if two customers have time overlap. For example, the arrival time of customer $u_1$ is earlier than the departure time of $u_2$, so $u_1$ and $u_2$ are not connected. On the other hand, the first customer and the third customer plan to use a taxi from 17:13 to 17:15. Therefore, $u_2$ and $u_3$ are connected, as shown in Fig. \ref{app} (b). After obtaining the graph description of the relationship between customers' schedules, the generated graph is optimized by GNN-1N with a specific color number. The color number is the number of available taxis, which is four here. Figure \ref{app} (c) shows the final color assignments with four colors after optimization, and no conflict is found in the solution. According to the color assignment, seven customers are assigned into four groups, that is, $\{u_1, u_2, u_7\}$, $\{u_3\}$, $\{u_4, u_6\}$, and $\{u_5\}$.

\begin{figure}[ht]
	
	\centering
	\subfigure[]{
		
		\includegraphics[width=0.7\linewidth]{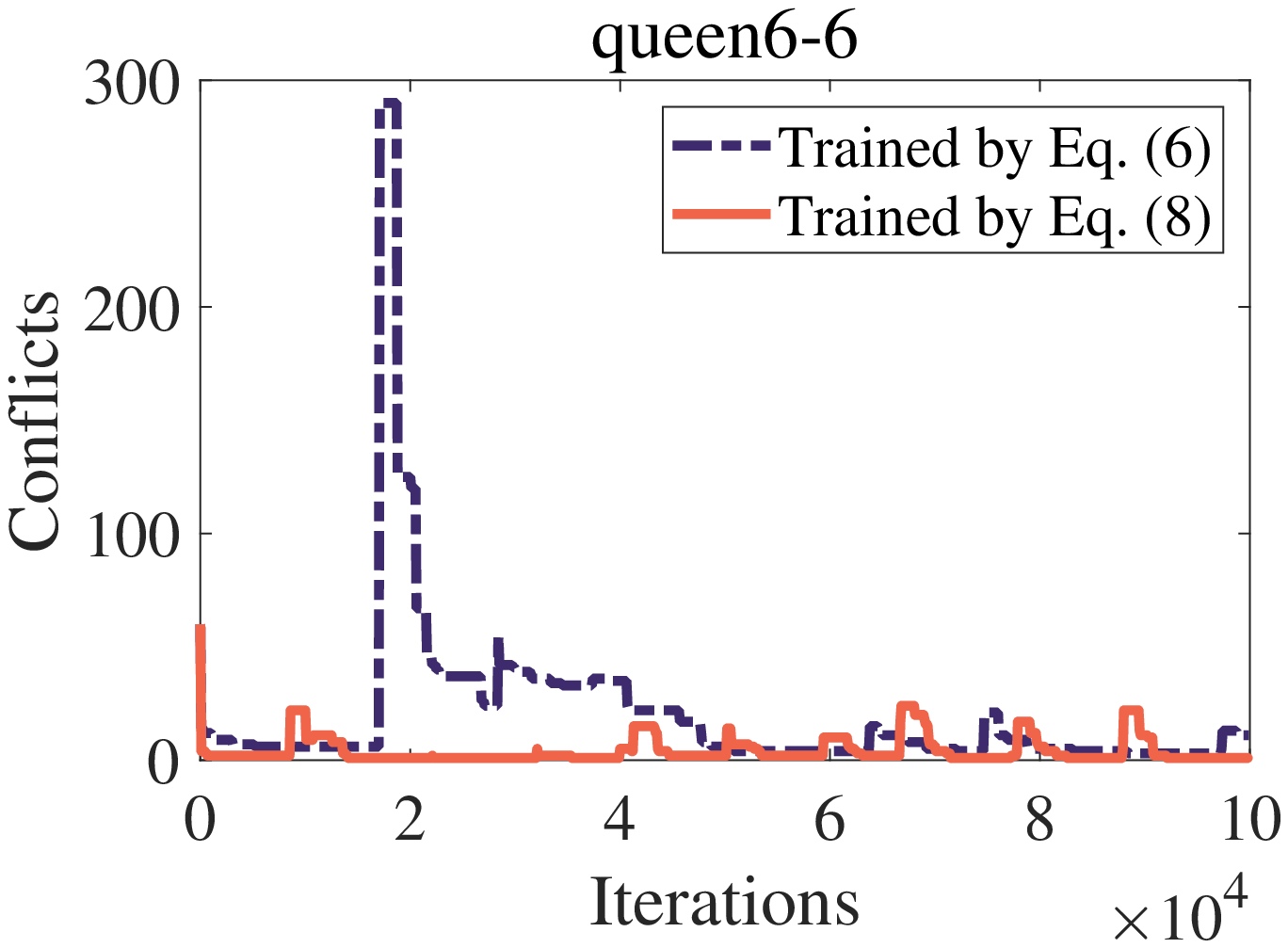}
		
		\label{ab-queen6}
	}%
	
	\subfigure[]{
		
		\includegraphics[width=0.7\linewidth]{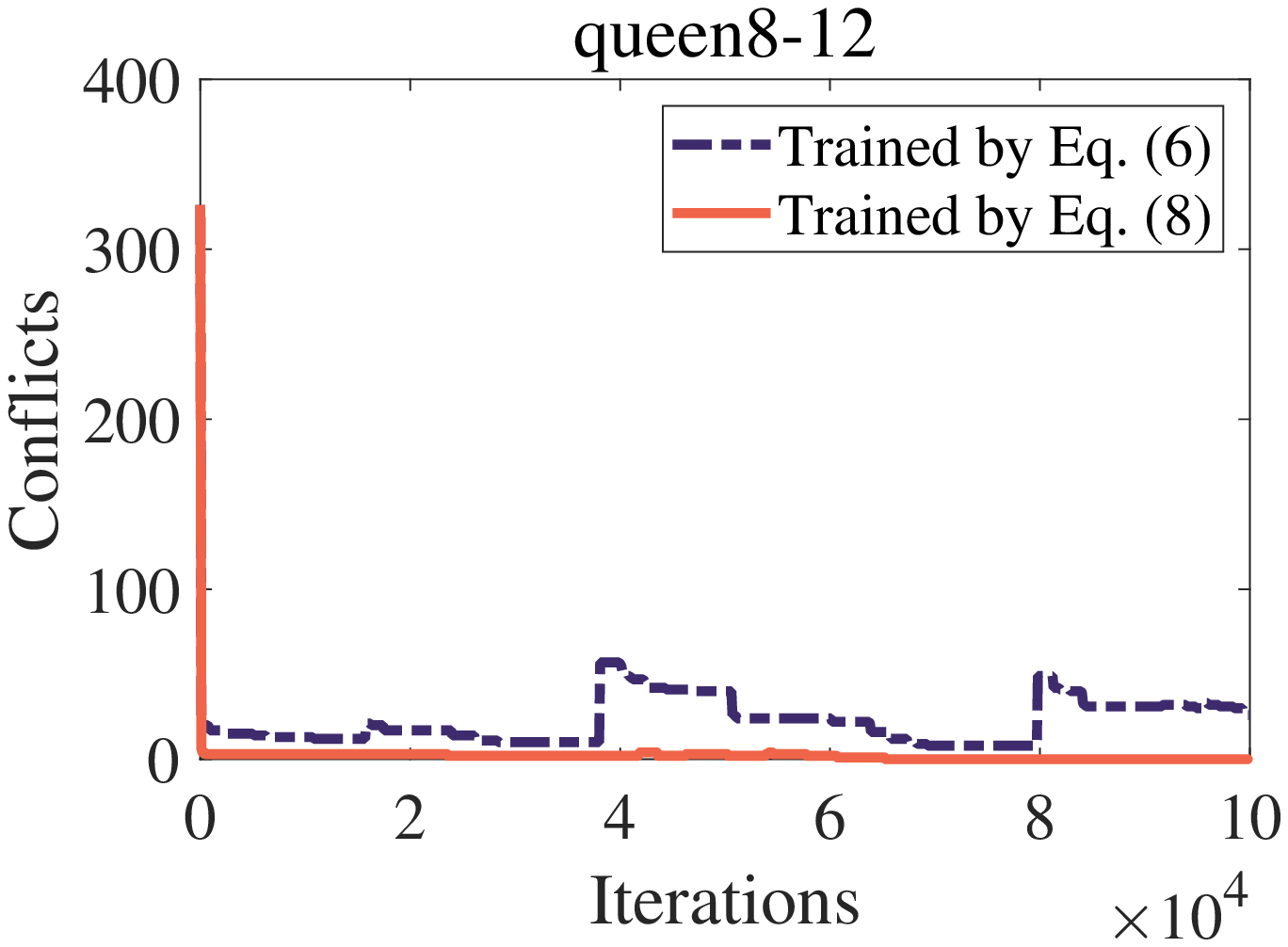}
		
		\label{ab-queen812}
	}%

	\caption{Convergence profiles of conflicts trained by Eq. (6) and Eq. (8) on queen6-6 and queen8-12. Eq. (6) contains $f_{utility}$ only, and Eq. (8) is summation of $f_{utility}$ and $\lambda f_{conv}$, where $f_{conv}$ is the self-information.}
 \label{Abolation}
\end{figure}
\begin{figure}[ht]
	
	\centering
	\includegraphics[width=1\linewidth]{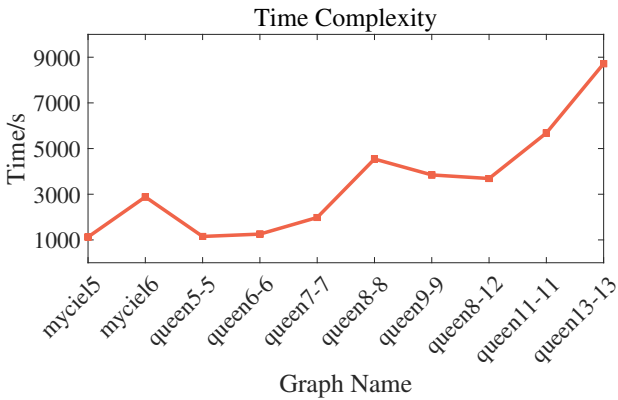}

	\caption{The runtimes (in seconds) of GNN-1N required on each graph coloring problem for $10^5$ iterations.}
 \label{time}
\end{figure}

\subsection{Ablation Study and Time Complexity Analysis}

An ablation study is conducted to show the effectiveness of self-information term $f_{conv}$ (Eq. (7)) included in the loss function $F$. Figure \ref{Abolation} shows the conflicts trained by the loss function with and without $f_{conv}$ on queen6-6 and queen8-12 over $10^5$ iterations. To be specific, Eq. (6) is the $f_{utility}$ proposed in \cite{PI-GNN}, and Eq. (8) is the loss function proposed in this paper with adding the self-information term. The hyperparameters are obtained directly from \cite{PI-GNN}, and the $\lambda$ in Eq. (8) is set to be $0.25$. As we can see in Fig. \ref{Abolation}, the conflicts curve obtained by Eq. (6) is unstable and fluctuates dramatically sometimes. By contrast, the conflicts curve trained by Eq. (8) decreases smoothly. The curves in Fig. \ref{Abolation} indicate the stabilization function of self-information, which can be attributed to the convergence term in Eq. (\ref{F}). 

The runtime (in seconds) is plotted in Fig. \ref{time}, which shows the time required by GNN-1N to run $10^5$ iterations on one graph coloring problem. It is reasonable that the runtimes increase with the number of nodes and edges increase, because the computational cost is mainly concentrated in the process of node aggregation and backpropagation. In general, the time complexity of GNN-1N is similar to PI-GNN \cite{PI-GNN}, as no additional computation is added to the proposed algorithm.

\section{Conclusion and Future Work}


The graph coloring problem is a classical graph-based problem aiming to find a color assignment using a given number of colors. As the GC problem is an NP-hard problem, it is almost impossible to obtain a feasible solution in an acceptable time. Moreover, due to its graph property, such problems cannot be easily and effectively solved by conventional neural networks. In this work, we propose an unsupervised graph neural network (GNN-1N) tailored for solving GCPs, which combines negative message passing with normal message passing to handle heterophily. Besides, a loss function with the utility-based objective and convergence-based objective is proposed for unsupervised learning. Experimental results on public datasets show that GNN-1N outperforms five state-of-the-art peer algorithms. In addition, a toy real-world application of graph coloring problems is also given to demonstrate further the effectiveness of GNN-1N.

Solving graph-based heterogeneous problems with GNNs is still in its infancy. The following three improvements could be made on the proposed model. First, we can consider pre-/post-processing to further decrease the number of conflicts. Second, the dynamic graph coloring problems are worthy of investigation, as the conditions may change in the real world. Finally, the fairness of color assignment is an interesting topic to examine. For example, each color should be used roughly the same number of times while making sure that there is no conflict in connecting nodes. For the taxi scheduling problem, each taxi should have a similar number of customers. Therefore, fairness coloring is of great practical importance. 



\bibliographystyle{IEEEtran}
\bibliography{IEEEabrv,mybib}


\end{document}